\documentclass[11pt]{article}

\usepackage{deauthor}
\usepackage{times,graphicx}
\usepackage[utf8]{inputenc} 
\usepackage[T1]{fontenc}    
\usepackage{url}            
\usepackage{booktabs}       
\usepackage{amsfonts}       
\usepackage{nicefrac}       
\usepackage{microtype}      
\usepackage{xcolor}         
\usepackage{graphicx}
\usepackage{caption}
\usepackage{enumitem}
\usepackage{subcaption}
\usepackage{caption}
\usepackage[subtle,margins=normal,leading=normal]{savetrees} 
\usepackage[driverfallback=dvipdfm]{hyperref}       

\begin{document}

\title{NVIDIA FLARE:\\ Federated Learning from Simulation to Real-World}

%


\author{%
  Holger R. Roth\quad
  Yan Cheng\quad
  Yuhong Wen\quad
  Isaac Yang\quad
  Ziyue Xu\quad
  Yuan-Ting Hsieh\quad\\
  Kristopher Kersten\quad
  Ahmed Harouni\quad 
  Can Zhao\quad
  Kevin Lu\quad
  Zhihong Zhang\quad
  Wenqi Li\quad\\
  Andriy Myronenko\quad
  Dong Yang\quad
  Sean Yang\quad
  Nicola Rieke\quad
  Abood Quraini\quad
  Chester Chen\quad\\
  Daguang Xu\quad
  Nic Ma\quad
  Prerna Dogra\quad
  Mona Flores\quad
  Andrew Feng\\
  \\
  NVIDIA Corporation\thanks{Contact: \texttt{\{hroth,yanc,chesterc,daguangx,pdogra,andyf\}@nvidia.com}}\\
  Shanghai, China\\ 
  Munich, Germany\\
  Bethesda, Santa Clara, USA\\
}



\maketitle

\setcounter{footnote}{0} 
\begin{abstract}
Federated learning (FL) enables building robust and generalizable AI models by leveraging diverse datasets from multiple collaborators without centralizing the data. We created NVIDIA FLARE\footnote{Code is available at \url{https://github.com/NVIDIA/NVFlare}.} as an open-source software development kit (SDK) to make it easier for data scientists to use FL in their research and real-world applications. 
The SDK includes solutions for state-of-the-art FL algorithms and federated machine learning approaches, which facilitate building workflows for distributed learning across enterprises and enable platform developers to create a secure, privacy-preserving offering for multiparty collaboration utilizing homomorphic encryption or differential privacy. 
The SDK is a lightweight, flexible, and scalable Python package. It allows researchers to apply their data science workflows in any training libraries (PyTorch, TensorFlow, XGBoost, or even NumPy) in real-world FL settings. This paper introduces the key design principles of NVFlare and illustrates some use cases (e.g., COVID analysis) with customizable FL workflows that implement different privacy-preserving algorithms.
\end{abstract}

\section{Introduction}
Federated learning (FL) has become a reality for many real-world applications~\cite{rieke2020future}. It enables multinational collaborations on a global scale to build more robust and generalizable machine learning and AI models. 
In this paper, we introduce NVIDIA FLARE (NVFlare), an open-source software development kit (SDK) that makes it easier for data scientists to collaborate to develop more generalizable and robust AI models by sharing model weights rather than private data.
While FL is attractive in many industries, it is particularly beneficial for healthcare applications where patient data needs to be protected. For example, FL has been used for predicting clinical outcomes in patients with COVID-19~\cite{dayan2021federated} or to segment brain lesions in magnetic resonance imaging~\cite{sheller2018multi,sheller2020federated}. NVFlare is not limited to applications in healthcare and is designed to allow cross-silo FL~\cite{kairouz2019advances} across enterprises for different industries and researchers. 

In recent years, several efforts (both open-source and commercial) have been made to bring FL technology into the healthcare sector and other industries, like TensorFlow Federated~\cite{abadi2016tensorflow}, PySyft~\cite{ziller2021pysyft}, FedML~\cite{he2020fedml}, FATE~\cite{liu2021fate}, Flower~\cite{beutel2020flower}, OpenFL~\cite{reina2021openfl}, Fed-BioMed~\cite{silva2020fed}, IBM Federated Learning~\cite{ludwig2020ibm}, HP Swarm Learning~\cite{warnat2021swarm}, FederatedScope~\cite{xie2022federatedscope}, FLUTE~\cite{dimitriadis2022flute}, and more. Some focus on simulated FL settings for researchers, while others prioritize production settings. NVFlare aims to be useful for both scenarios: 1) for researchers by providing efficient and extensible simulation tools and 2) by providing an easy path to transfer research into real-world production settings, supporting high availability and server failover, and by providing additional productivity tools such as multi-tasking and admin commands.

\section{NVIDIA FLARE Overview}

NVIDIA FLARE -- or short NVFlare -- stands for ``\textbf{NV}IDIA \textbf{F}ederated \textbf{L}earning \textbf{A}pplication \textbf{R}untime \textbf{E}nvironment''. 
The SDK enables researchers and data scientists to adapt their machine learning and deep learning workflows to a federated paradigm. It enables platform developers to build a secure, privacy-preserving offering for distributed multiparty collaboration.

NVFlare is a lightweight, flexible, and scalable FL framework implemented in Python that is agnostic to the underlying training library. Developers can bring their own data science workflows implemented in PyTorch, TensorFlow, or even in pure NumPy, and apply them in a federated setting.
A typical FL workflow such as the popular federated averaging (FedAvg) algorithm~\cite{mcmahan2017communication}, can be implemented in NVFlare using the following main steps. Starting from an initial global model, each FL client trains the model on their local data for a while and sends model updates to the server for aggregation. The server then uses the aggregated updates to update the global model for the next round of training. This process is iterated many times until the model converges.

Though used heavily for federated deep learning, NVFlare is a generic approach for supporting collaborative computing across multiple clients. NVFlare provides the \textit{Controller} programming API for researchers to create workflows for coordinating clients for collaboration. FedAvg is one such workflow. Another example is cyclic weight transfer~\cite{chang2018distributed}. 
The central concept of collaboration is the notion of ``task''. An FL controller assigns tasks (e.g., deep-learning training with model weights) to one or more FL clients and processes results returned from clients (e.g., model weight updates). The controller may assign additional tasks to clients based on the processed results and other factors (e.g., a pre-configured number of training rounds). This task-based interaction continues until the objectives of the study are achieved. 
%
\begin{figure}[htbp]
    \centering
    \includegraphics[width=0.45\textwidth]{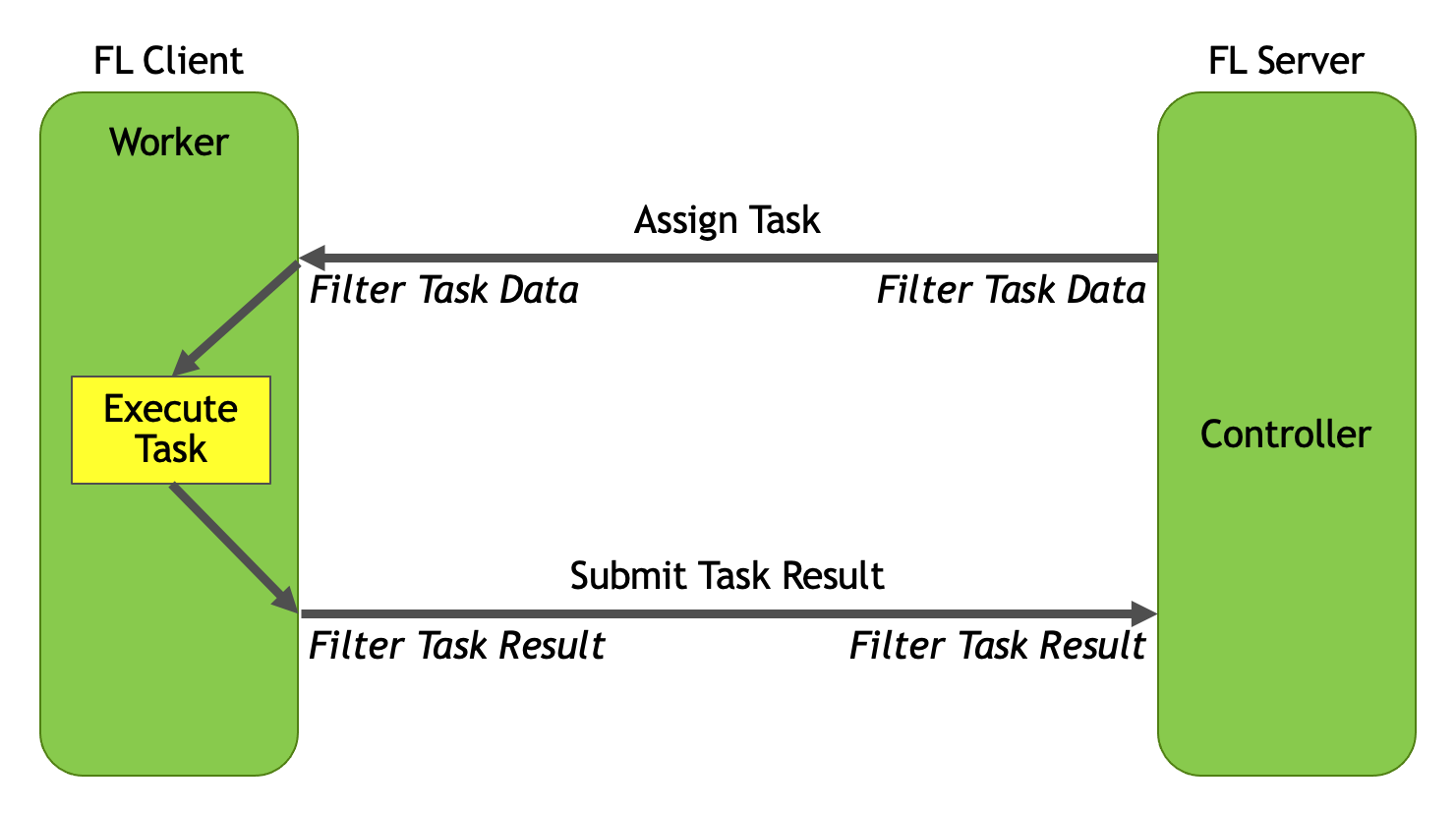}
    \caption{NVFlare job execution. The \textit{Controller} is a Python object that controls or coordinates the \textit{Workers} to get a job done. The controller is run on the FL server. A \textit{Worker} is capable of performing tasks. \textit{Workers} run on FL clients. \label{fig:job}}
\end{figure}
%
The API supports typical controller-client interaction patterns like broadcasting a task to multiple clients, sending a task to one or more specified clients, or relaying a task to multiple clients sequentially. Each interaction pattern has two flavors: wait (block until client results are received) or no-wait. A workflow developer can use these interaction patterns to create innovative workflows. For example, the \textit{ScatterAndGather} controller (typically used for FedAvg-like algorithms) is implemented with the \textit{broadcast\_and\_wait} pattern, and the \textit{CyclicController} is implemented with the \textit{relay\_and\_wait pattern}. The controller API allows the researcher to focus on the control logic without needing to deal with underlying communication issues. Figure~\ref{fig:job} shows the principle.
Each FL client acts as a worker that simply executes tasks assigned to it (e.g., model training) and returns execution results to the controller. At each task interaction, there can be optional filters that process the task data or results before passing it to the \textit{Controller} (on the server side) or task executor (client side). The filter mechanism can be used for data privacy protection (e.g., homomorphic encryption/decryption or differential privacy) without having to alter the training algorithms.

\paragraph{Key Components}
NVFlare is built on a componentized architecture that allows FL workloads to move from research and simulation to real-world production deployment. Some of the key components of this SDK include:

\begin{itemize}
    \item \textbf{FL Simulator} for rapid development and prototyping.
    \item \textbf{NVFlare Dashboard} for simplified project management, secure provisioning, and deployment, orchestration.
    \item \textbf{Reference FL algorithms} (e.g., FedAvg, FedProx, SCAFFOLD) and workflows, like scatter and gather, cyclic, etc.
    \item \textbf{Privacy preservation} with differential privacy, homomorphic encryption, and more.
    \item \textbf{Specification-based API} for extensibility, allowing customization with plug-able components.
    \item \textbf{Tight integration} with other learning frameworks like MONAI~\cite{cardoso2022monai}, XGBoost~\cite{Chen:2016:XST:2939672.2939785}, and more.    
\end{itemize}

\paragraph{High-Level Architecture} NVFlare is designed with the idea that less is more, using a specification-based design principle to focus on what is essential.
This allows other people to be able to do what they want to do in real-world applications by following clear API definitions. FL is an open-ended space. The API-based design allows others to bring their implementations and solutions for various components. Controllers, task executors, and filters are just examples of such extensible components. 
NVFlare provides an end-to-end operation environment for different personas. It provides a comprehensive provisioning system that creates security credentials for secure communications to enable the easy and secure deployment of FL applications in the real world. It also provides an FL Simulator for running proof-of-concept studies locally.
In production mode, the researcher conducts an FL study by submitting jobs using admin commands using Notebooks or the NVFlare Console -- an interactive command tool. NVFlare provides many commands for system operation and job management. With these commands, one can start and stop a specific client or the entire system, submit new jobs, check the status of jobs, create a job by cloning from an existing one, and much more.

With NVFlare's component-based design, a job is just a configuration of components needed for the study. For the control logic, the job specifies the controller component to be used and any components required by the controller. 
\section{System Concepts}
A NVFlare system is a typical client-server communication system that comprises one or more FL server(s), one or more FL client(s), and one or more admin clients. The FL Servers open two ports for communication with FL clients and admin clients. FL clients and admin clients connect to the opened ports. FL clients and admin clients do not open any ports and do not directly communicate with each other.
The following is an overview of the key concepts and objects available in NVFlare and the information that can be passed between them.

\paragraph{Workers and Controller} NVFlare’s collaborative computing is achieved through the \textit{Controller}/\textit{Worker} interactions. 

\paragraph{Shareable} Object that represents a communication between server and client. Technically, the \textit{Shareable} is implemented as a Python dictionary that could contain different information, e.g., model weights.

\paragraph{Data Exchange Object (DXO)} Standardizes the data passed between the communicating parties. One can think of the \textit{Shareable} as the envelope and the \textit{DXO} as the letter. Together, they comprise a message to be shared between communicating parties.

\paragraph{FLComponent} The base class of all the FL components. Executors, controllers, filters, aggregators, and their subtypes are all \textit{FLComponents}.
\textit{FLComponent} comes with some useful built-in methods for logging, event handling, auditing, and error handling.

\paragraph{Executors} Type of \textit{FLComponent} for FL clients that has an execute method that produces a \textit{Shareable} from an input \textit{Shareable}. NVFlare provides both single- and multi-process executors to implement different computing workloads.

\paragraph{FLContext} One of the most important features of NVFlare is to pass data between the FL components. \textit{FLContext} is available to every method of all common FLComponent types. Through \textit{FLContext}, the component developer can get services provided by the underlying infrastructure and share data with other components of the FL system.

\paragraph{Communication Drivers}
NVFlare abstracts the communication layers out so that different deployment scenarios can implement customizable communication drivers. By default, we use GRPC for data communication in task-based communication. However, the driver can be replaced to run other communication protocols, for example, TCP. The customizable nature of communication in NVFlare allows for both server-centric and peer-to-peer communication patterns. This enables the user to utilize both scatter and gather-type workflows like FedAvg~\cite{mcmahan2017communication}, decentralized training patterns like swarm learning~\cite{warnat2021swarm}, or direct peer-to-peer communication as in split learning~\cite{gupta2018distributed}.

\begin{figure}[htbp]
    \centering
    \includegraphics[width=0.75\textwidth]{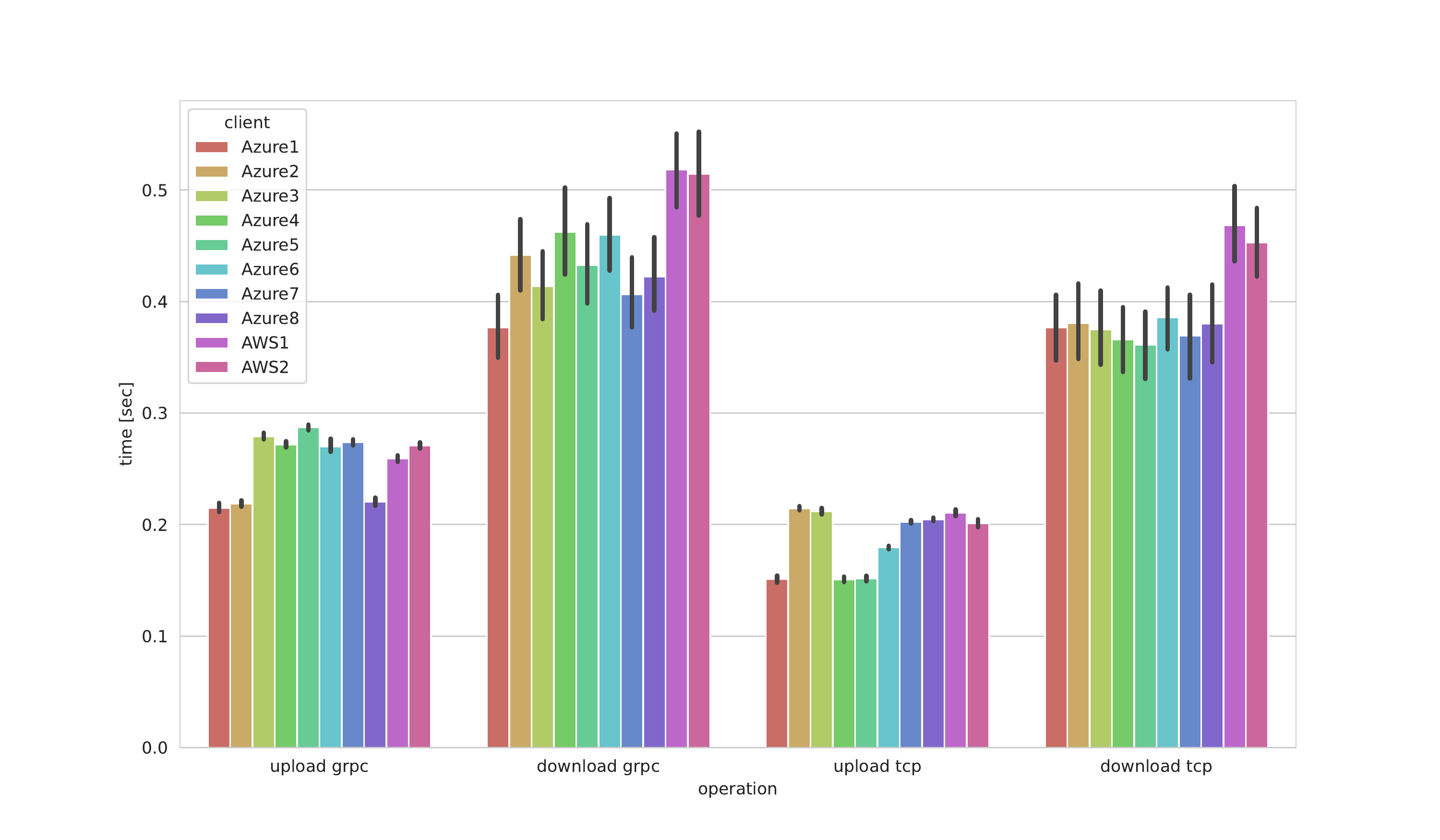}
    \caption{Comparison of GRPC and TCP communication drivers in NVFlare. The server is running on Azure. The clients are distributed between Azure and AWS. The message size is $\sim$18MB. Communication times were measured over 100 rounds of FedAvg. Error bars indicate the 95\% confidence intervals. \label{fig:cloud_comm}}
\end{figure}

Fig.~\ref{fig:cloud_comm} compares the times for model upload and download from the client's perspective using different communication protocols available in NVFlare using a model of $\sim$18MB in size.

The experiment runs in a multi-cloud environment with the server and eight clients running on Azure, while two clients run on AWS. One can observe that the global model download is slower as all clients are trying to download the global model at the same time, and hence the server is more busy. In contrast, the clients' model uploads happen at slightly different times and therefore are faster. One can also see how this multi-cloud setup causes the clients on AWS to take slightly longer during model download due to communication across different cloud infrastructures. 

\paragraph{Filters} Filters in NVFlare are a type of \textit{FLComponent} that have a process method to transform the \textit{Shareable} object between the communicating parties. A Filter can provide additional processing to shareable data before sending or after receiving from a peer. Filters can convert data formats and a lot more and are NVFlare's primary mechanism for data privacy protection~\cite{li2019privacy,hatamizadeh2022gradient}:
\begin{itemize}
\item \textit{ExcludeVars} to exclude variables from shareable.
\item \textit{PercentilePrivacy} for truncation of weights by percentile.
\item \textit{SVTPrivacy} for differential privacy through sparse vector techniques.
\item Homomorphic encryption filters used for secure aggregation.
\end{itemize}
As an example, we show the average encryption, decryption, and upload times when using homomorphic encryption for secure aggregation\footnote{\url{https://developer.nvidia.com/blog/federated-learning-with-homomorphic-encryption}}. We compare raw data to encrypted model gradients uploaded in Table~\ref{tab:he} when hosting the server on AWS\footnote{For reference, we used an \href{https://aws.amazon.com/ec2/instance-types}{m5a.2xlarge} instance with eight vCPUs, 32-GB memory, and up to 2,880 Gbps network bandwidth.} and connecting 30 client instances using an on-premise GPU cluster. One can see the longer upload times due to the larger message sizes needed by homomorphic encryption.
\begin{table}[htbp]
\centering
\captionsetup{width=.45\textwidth}
\caption{Federated learning exchanging homomorphic encrypted vs. raw model updates. \label{tab:he}}
\footnotesize
 \begin{tabular}{||l r r||} 
 \hline
\textbf{Time in seconds}	& \textbf{Mean} & \textbf{Std. Dev.} \\ [0.5ex] 
 \hline\hline
Encryption & 5.01 & 1.18 \\
Decryption & 0.95 & 0.04 \\
Enc. upload & 38.00 & 71.17 \\
Raw upload & 21.57 & 74.23 \\ [1ex] 
 \hline
 \end{tabular}
\end{table}
%
\paragraph{Event Mechanism} NVFlare comes with a powerful event mechanism that allows dynamic notifications to be sent to all event handlers. This mechanism enables data-based communication among decoupled components: one component fires an event when a certain condition occurs, and other components can listen to that event and processes the event data. Each \textit{FLComponent} is automatically an event handler. To listen to and process an event, one can simply implement the \textit{handle\_event()} method and process desired event types. Events represent some important moments during the execution of the system logic. For example, before and after aggregation or when important data becomes available, e.g., a new ``best'' model was selected.
\subsection{Productivity Features}
NVFlare contains features that enable efficient, collaborative, and robust computing workflows.

\paragraph{Multi-tasking} For systems with a large capacity, computing resources could be idle most of the time. NVFlare implements a resource-based multi-tasking solution, where multiple jobs can be run concurrently when overall system resources are available. 
Multi-tasking is made possible by a job scheduler on the server side that constantly tries to schedule a new job.
For each job to be scheduled, the scheduler asks each client whether they can satisfy the required resources of the job (e.g., number of GPU devices) by querying the client's resource manager. If all clients can meet the requirement, the job will be scheduled and deployed to the clients.

\paragraph{High Availability and Server Failover} To avoid the FL server as a single point of failure, a solution has been implemented to support multiple FL servers with automatic cut-over when the currently active server becomes unavailable. 
Therefore, a component called \textit{Overseer} is added to facilitate automatic cut-over. The \textit{Overseer} provides the authoritative endpoint info of the active FL server. All other system entities (FL servers, FL clients, admin clients) constantly communicate (i.e., every 5 seconds) with the Overseer to obtain and act on such information. 
If the server cutover happens during the execution of a job, then the job will continue to run on the new server. Depending on how the controller is written, the job may or may not need to restart from the beginning but can continue from a previously saved snapshot. 

\paragraph{Simulator} NVFlare provides a simulator to allow data scientists and system developers to easily write new \textit{FLComponents} and novel workflows.
The simulator is a command line tool to run a NVFlare job. To allow simple experimentation and debugging, the FL server and multiple clients run in the same process during simulation. A multi-process option allows efficient use of resources, e.g., training multiple clients on different GPUs. The simulator follows the same job execution as in real-world NVFlare deployment. Therefore, components developed in simulation can be directly deployed in real-world federated scenarios.

\subsection{Secure Provisioning in NVFlare}
Security is an important requirement for FL systems. NVFlare provides security solutions in the following areas: authentication, communication confidentiality, user authorization, data privacy protection, auditing, and local client policies.
\begin{figure}[htbp]
    \centering
    \includegraphics[width=0.6\textwidth]{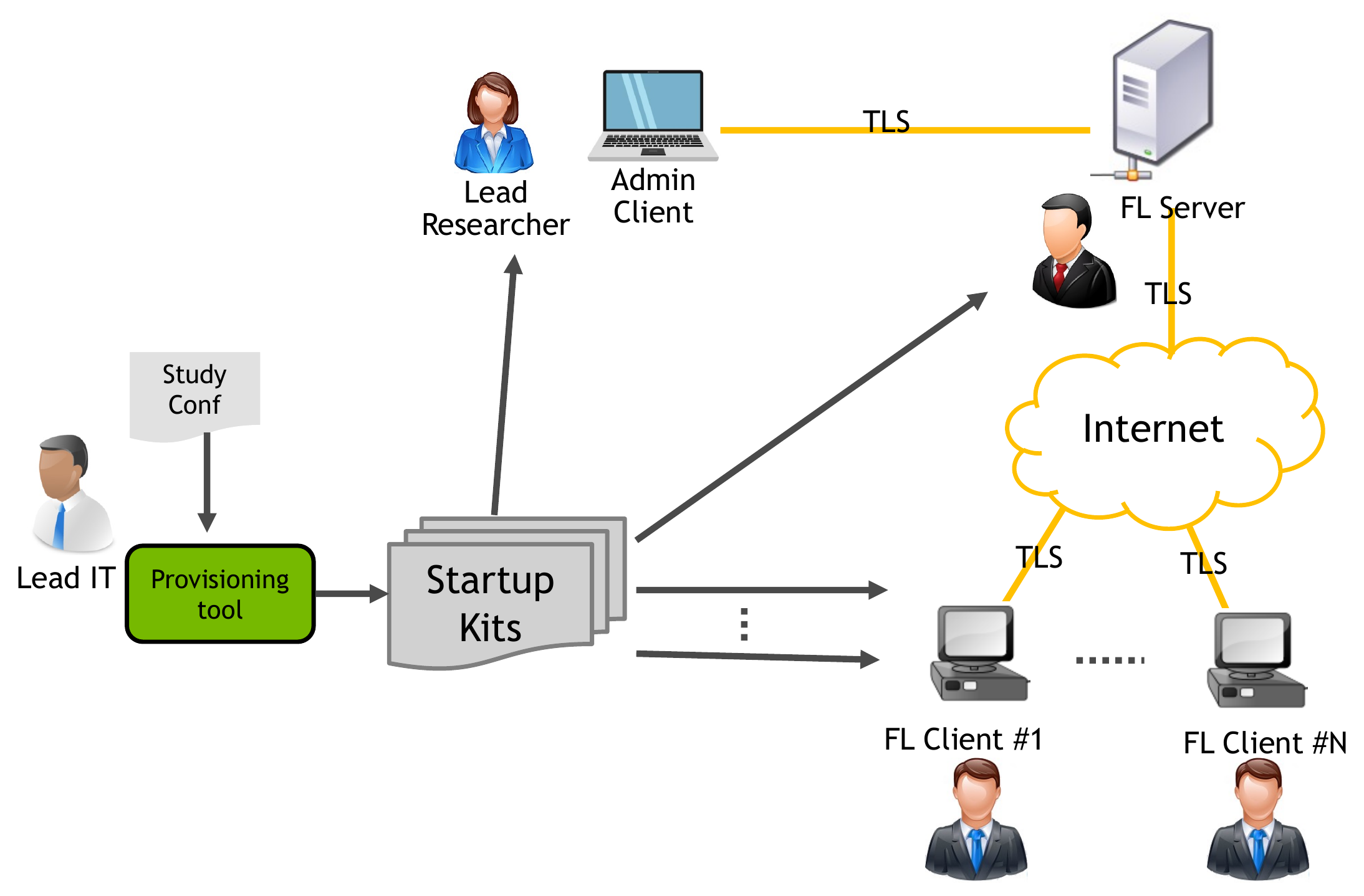}
    \caption{High-level steps for running a real-world study with secure provisioning with NVFlare. \label{fig:provisioning}}
\end{figure}
\paragraph{Authentication} NVFlare ensures the identities of communicating peers using mutual Transport Layer Security (TLS). Each participating party (FL Servers, Overseer, FL Clients, Admin Clients) must be properly provisioned. Once provisioned, each party receives a startup kit containing TLS credentials (public cert of the root, the party's own private key and certificate) and system endpoint information, see Fig.~\ref{fig:provisioning}. Each party can only connect to the NVFlare system with the startup kit.
Communication confidentiality is also achieved with the use of TLS-based messaging.

\paragraph{Federated Authorization} NVFlare's admin command system is very rich and powerful. Not every command is for everyone. NVFlare implements a role-based user authorization system that controls what a user can or cannot do. At the time of provision, each user is assigned a role. Authorization policies specify which commands are permitted for which roles.
Each FL client can define its authorization policy that specifies what a role can or cannot do to the client. For example, one client could allow a role to run jobs from any researchers. In contrast, another client may only allow jobs submitted by its researchers (i.e., the client and the job submitter belong to the same organization).

NVFlare automatically records all user commands and job events in system audit files on both the server and client sides. In addition, the audit API can be used by application developers to record additional events in the audit files.

\paragraph{Client-Privacy} NVFlare enhances the overall system security by allowing each client to define its policies for authorization, data privacy (filters), and computing resource management. The client can change its policies at any time after the system is up and running without having to be re-provisioned. For example, the client could require all jobs running on it to be subject to a set of filters. The client could also change the number of computing resources (e.g., GPU devices) to be used by the FL client.

\section{Federated Data Science}
As a general distributed computing platform, NVFlare can be used for various applications in different industries. Here we describe some of the most common use cases where NVFlare was deployed.

\subsection{Federated Deep Learning} 
\begin{figure}[htbp]
\centering
\begin{subfigure}[t]{0.45\textwidth}
    \includegraphics[width=\textwidth]{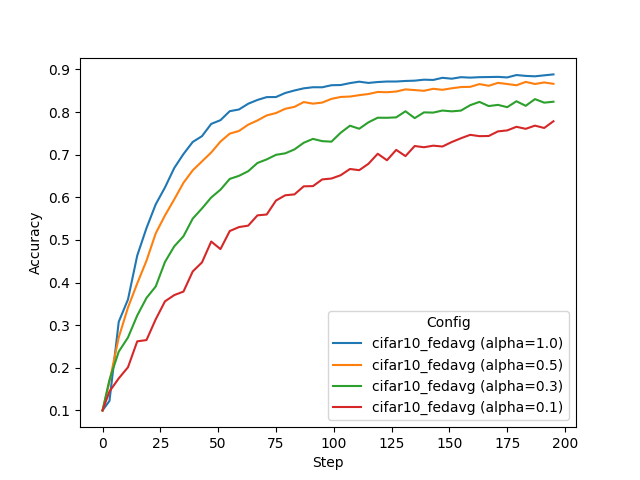}
    \caption{\footnotesize FedAvg with increasing levels of heterogeneity (smaller $\alpha$ values). \label{fig:fl_alpha}}
\end{subfigure}\qquad
\begin{subfigure}[t]{0.45\textwidth}
    \includegraphics[width=\textwidth]{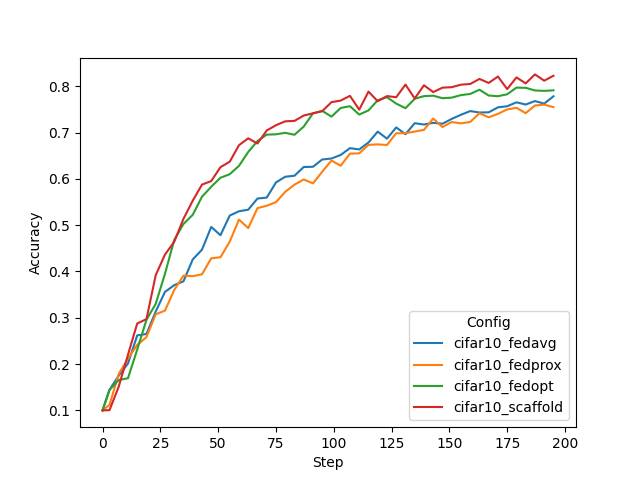}
    \caption{\footnotesize FL algorithms with a heterogeneous data split ($\alpha$=0.1). \label{fig:fl_algos}}
\end{subfigure}
\caption{Federated learning experiments with NVFlare. \label{fig:fl}}
\end{figure}

A go-to example dataset for benchmarking different FL algorithms is CIFAR-10~\cite{krizhevsky2009learning}. NVFlare allows users to experiment with different algorithms and data splits using different levels of heterogeneity based on a Dirichlet sampling strategy~\cite{wang2020federated}. Figure~\ref{fig:fl_alpha} shows the impact of varying alpha values, where lower values cause higher heterogeneity on the performance of the FedAvg.

Apart from FedAvg, currently available in NVFlare include FedProx~\cite{li2020federated}, FedOpt~\cite{reddi2020adaptive}, and SCAFFOLD~\cite{karimireddy2020scaffold}. Figure~\ref{fig:fl_algos} compares an $\alpha$ setting of 0.1, causing a high data heterogeneity across clients and its impact on more advanced FL algorithms, namely FedProx, FedOpt, and SCAFFOLD. 
FedOpt and SCAFFOLD show markedly better convergence rates and achieve better performance than FedAvg and FedProx with the same alpha setting. 
SCAFFOLD achieves this by adding a correction term when updating the client models, while FedOpt utilizes SGD with momentum to update the global model on the server. Therefore, both perform better with the same number of training steps as FedAvg and FedProx.

Other algorithms available in or coming soon to NVFlare include federated XGBoost~\cite{Chen:2016:XST:2939672.2939785}, Ditto~\cite{li2021ditto}, FedSM~\cite{xu2022closing}, Auto-FedRL~\cite{guo2022auto}, and more.

\subsection{Federated Machine Learning} 
Traditional machine learning methods, such as linear models, support vector machine (SVM), and k-means clustering, can be formulated under a federated setting.

With certain libraries, the federated machine learning algorithms need to be designed considering two factors: algorithm-wise, each of these models has distinct training schemes and model representations; and implementation-wise, popular libraries providing these functionalities (e.g., scikit-learn, XGBoost) have different APIs and inner logics. Hence, when developing an FL variant of a particular traditional machine learning method, several questions need to be answered at these two levels: 

First, at the algorithm level, we need to break down the optimization process into individual steps/rounds (if possible) and have answers to three major questions:
\begin{enumerate}
    \item What information should clients share with the server?
    \item How should the server aggregate the collected information from clients?
    \item What should clients do with the global aggregated information received from the server?
\end{enumerate}

Second, at the implementation level, we need to know what APIs are available and how to utilize them in a federated pipeline to implement a distributed version of the algorithm.

A major difference between federated traditional machine learning and federated deep learning is that, for traditional machine learning methods, the boundary between ``federated'' and ``distributed'', or even ``ensemble'', can be much more vague than for deep learning. Due to the characteristics of a given algorithm and its API design, the concepts can be equivalent. Take XGBoost and SVM, for example:
Algorithm-wise, XGBoost can distribute the training samples to several workers and construct trees based on the collected histograms from each worker. Such a process can be directly adopted under a federated setting because the communication cost is affordable. In this case, ``federated'' is equivalent to ``distributed'' learning.
API-wise, some algorithms can be constrained by their implementation. Take scikit-learn's SVM for instance. Although theoretically SVM can be formulated as an iterative optimization process, the API only supports one-shot ``fitting'' without the capability of separately calling the optimization steps. Hence a federated SVM algorithm using the scikit-learn library can only be implemented as a two-step process. In this case, ``federated'' is equivalent to ``ensemble''.

\begin{figure}[htbp]
    \centering
    \includegraphics[width=0.8\textwidth]{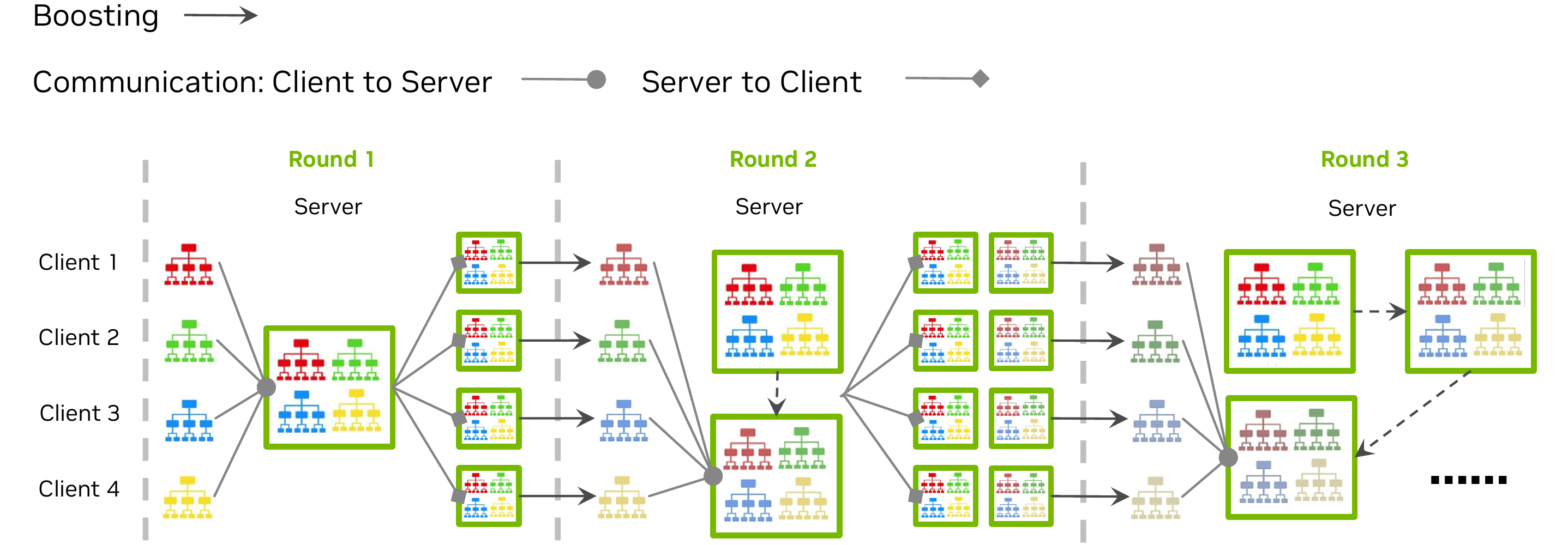}
    \caption{Tree-based federated XGBoost: a ``boosting of forests.'' \label{fig:tree_xgboost}}
\end{figure}

For clarification, we provide the full formulation for tree-based federated XGBoost, illustrated in Fig.~\ref{fig:tree_xgboost}:
\begin{enumerate}
\item XGBoost, by definition, is a sequential optimization process: each step adds one extra tree to the model to reduce the residual error. Hence, federated XGBoost can be formulated as follows: each round of FL corresponds to one boosting step at the local level. Clients share the newly added tree trained on local data with the server at the end of local boosting.
\item The model representation is a decision/regression tree. To aggregate the information from all clients, the server will bag all received trees to form a ``forest'' to be added to the global boosting model.
\item With the updated global model from the server, each client will continue the boosting process by learning a new tree starting from the global model of the boosted forest.
\end{enumerate}

\subsection{Split learning}
Split learning assumes a vertical data partitioning~\cite{yang2019federated} that can be useful in many distributed learning scenarios involving neural network architectures~\cite{gupta2018distributed}.

\begin{figure}[htbp]
\centering
\includegraphics[width=0.25\textwidth]{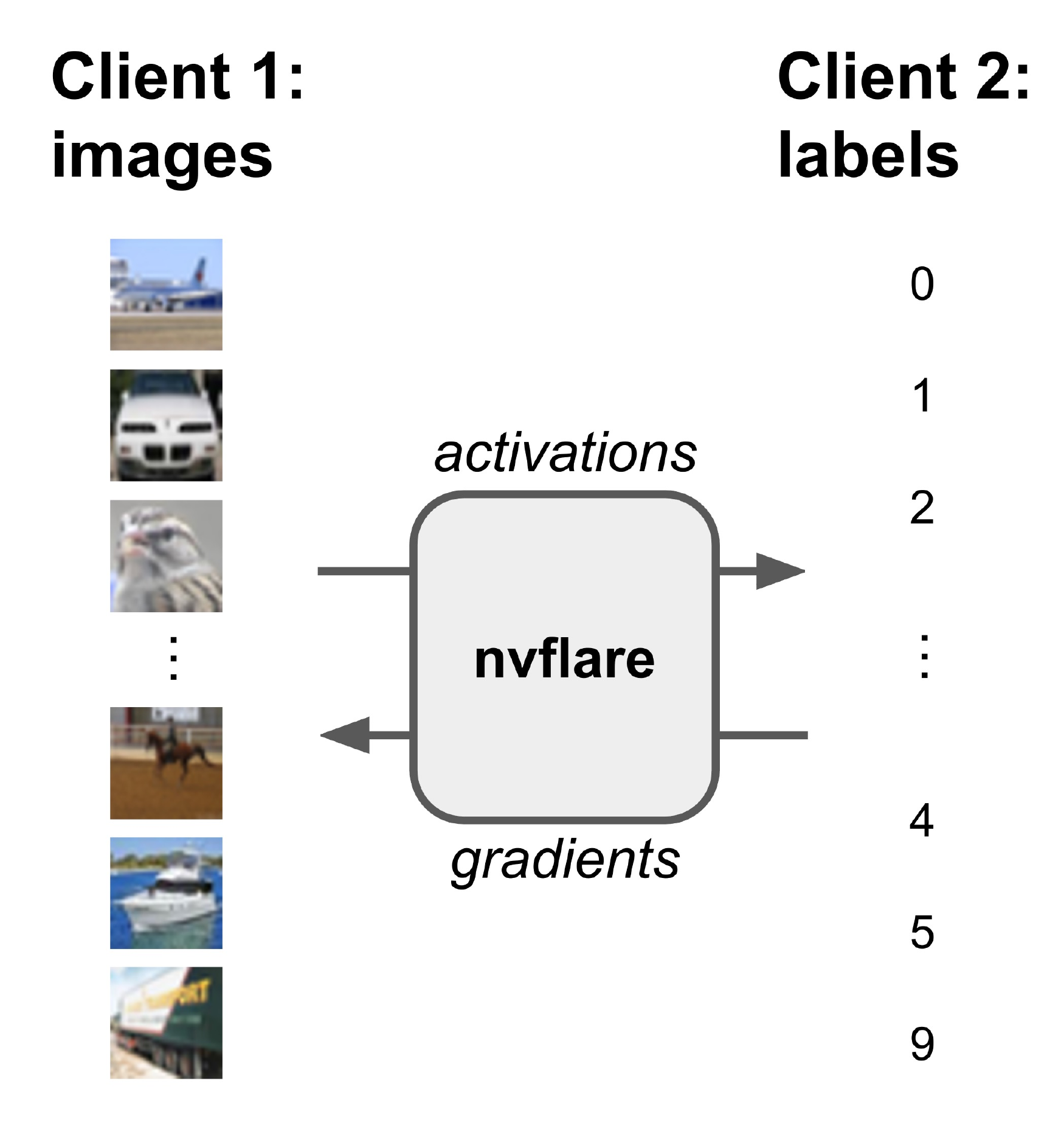}
\hspace{2em}
\footnotesize
\begin{tabular}[b]{||l r||} 
\hline
\textbf{Setup} & \textbf{Training Time [min]} \\ [0.5ex] 
\hline\hline
Simulated PyTorch & 19 \\
Routing through server (TCP) & 27  \\
Peer-to-peer (TCP) & 25  \\ [1ex] 
\hline
\end{tabular}
\caption{Simple split learning scenario using CIFAR-10. The table compares multiple communication patterns. Using 50,000 training samples and 15,625 rounds of communication with a batch size of 64. \label{fig:split_learning}} 
\end{figure}

As an introductory example, we can assume that one client holds the images, and the other holds the labels to compute losses and accuracy metrics. Activations and corresponding gradients are being exchanged between the clients using NVFlare, as illustrated in Fig.~\ref{fig:split_learning}. 
%
%
We use a cryptographic technique called private set intersection (PSI)~\cite{enwiki:1131516757} to compute the alignment between images and labels on both clients. NVFlare's implementation of PSI can be extended to multiple parties and applied to other use cases than split learning, e.g., requiring a secure and privacy-preserving alignment of different databases.

Using NVFlare's capability to implement different communication patterns, we can investigate the communication speed-ups one can achieve by implementing split learning using direct peer-to-peer communication as opposed to routing the messages between the two clients through a central server. 

The table in Fig.~\ref{fig:split_learning} compares the training speeds of split learning on the CIFAR-10 dataset in a local simulation scenario. First, we use the same PyTorch script to simulate split learning. Then, we implement two distributed solutions using NVFlare. One that routes the messages through the server and one using a direct peer-to-peer connection between the clients. As expected, the direct peer-to-peer connection is more efficient, achieving only a slight overhead in total training time compared to the standalone PyTorch script, which could not be translated to real-world scenarios.


\subsection{Federated Statistics} 
\begin{figure}[htbp]
\centering
\begin{subfigure}[c]{0.55\textwidth}
    \includegraphics[width=\textwidth]{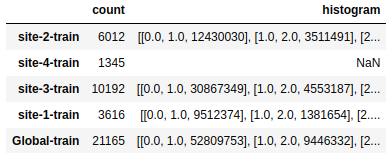}
    \caption{\footnotesize Federated statistics. Note the data of ``site-4'' violates the client's privacy policy and therefore does not share its statistics with the server. \label{fig:fedstats_table}}
\end{subfigure}\qquad
\begin{subfigure}[c]{0.35\textwidth}
    \includegraphics[width=\textwidth]{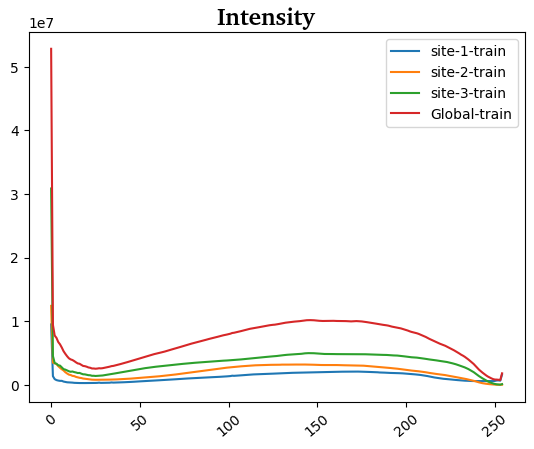}
    \caption{\footnotesize Histogram visualization. \label{fig:fedstats_histo}}
\end{subfigure}
\caption{Federated statistics with NVFlare. \label{fig:fedstats}}
\end{figure}

NVFlare provides built-in federated statistics operators (\textit{Controller} and \textit{Executors}) that will generate global statistics based on local client statistics. 
Each client could have one or more datasets, such as ``train'' and ``test'' datasets. Each dataset may have many features.
NVFlare will calculate and combine the statistics for each feature in the dataset to produce global statistics for all the numeric features. The output gathered on the server will be the complete statistics for all datasets in clients and global, as illustrated in Fig.~\ref{fig:fedstats}.
%

\section{Real-world Use Cases} 
NVFlare and its predecessors have been used in several real-world studies exploring FL for healthcare scenarios. 
The collaborations between multinational institutions tested and validated the utility of federated learning, pushing the
envelope for training robust, generalizable AI models. These initiatives included FL for breast mammography classification~\cite{roth2020federated}, prostate segmentation~\cite{sarma2021federated}, pancreas segmentation~\cite{wang2020federated}, and most recently, chest X-ray (CXR) and electronic health record (EHR) analysis to predict the oxygen requirement for patients arriving in the emergency department with symptoms of COVID-19~\cite{dayan2021federated}.
\begin{figure}[htbp]
\centering
\begin{subfigure}[t]{0.2\textwidth}
    \includegraphics[clip, trim={0.5em 0 0.5em 0}, height=7.5em]{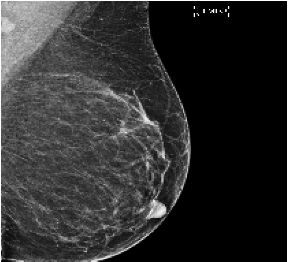}
    \caption{\footnotesize Mammography. \label{fig:breast}}
\end{subfigure}\hfill
\begin{subfigure}[t]{0.2\textwidth}
    \includegraphics[clip, trim={1em 0 1em 0}, height=7.5em]{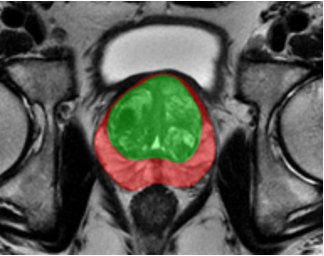}
    \caption{\footnotesize Prostate. \label{fig:prostate}}
\end{subfigure}\hfill
\begin{subfigure}[t]{0.2\textwidth}
    \includegraphics[clip, trim={1em 0 1em 0}, height=7.5em]{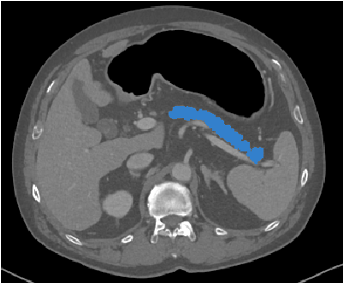}
    \caption{\footnotesize Pancreas. \label{fig:pancreas}}
\end{subfigure}\hfill
\begin{subfigure}[t]{0.2\textwidth}
    \includegraphics[clip, trim={0.5em 0 0.5em 0}, height=7.5em]{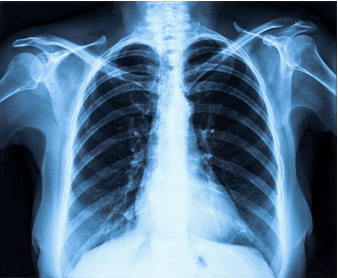}
    \caption{\footnotesize CXR \& EHR. \label{fig:cxr}}
\end{subfigure}
\caption{Real-world use cases of NVFlare. \label{fig:use_cases}}
\end{figure}
\section{Summary \& Conclusion}
We described NVFlare, an open-source SDK to make it easier for data scientists to use FL in their research and to allow an easy transition from research to real-world deployment.
As discussed above, NVFlare's \textit{Controller} programming API supports various interaction patterns between the server and clients over internet connections, which could be unstable. 
Therefore, the API design mitigates various failure conditions and unexpected crashes of the client machines, such as allowing developers to process timeout conditions properly.

NVFLare's unique flexibility and agnostic approach towards the deployed training libraries make it the perfect solution for integrating with different deep learning frameworks, including popular ones used for training large language models (LLM). With our dedication to addressing the current limitations of communication protocols, we are working towards supporting the communication of large message sizes, enabling the federated fine-tuning of AI models with billions of parameters, such as those used for ChatGPT~\cite{ouyang2022training} and GPT-4~\cite{openai2023gpt4}.
Moreover, our team is implementing parameter-efficient federated methods to adapt LLM models to downstream tasks~\cite{zhao2022reduce}, utilizing techniques such as prompt tuning~\cite{lester2021power} and p-tuning~\cite{liu2021gpt}, adapters~\cite{houlsby2019parameter,he2021towards}, LoRA~\cite{hu2021lora}, showing promising performance. Our commitment to innovation and excellence in this field ensures that we continue to push the boundaries of what is possible with federated learning.

We did not go into all details of exciting features available in NVFlare, like homomorphic encryption, TensorBoard streaming, provisioning web dashboard, integration with MONAI\footnote{\url{https://monai.io}}~\cite{monai2022github,cardoso2022monai}, etc. However, we hope that this overview of NVFlare gives a good starting point for developers and researchers on their journey to using FL and federated data science in simulation and the real world. 

NVFlare is an open-source project. We invite the community to contribute and grow NVFlare. For more information, please visit the code repository at \url{https://github.com/NVIDIA/NVFlare}.

\small
\bibliographystyle{abbrv}
\bibliography{ref}





\end{document}